\pdfoutput=1
\documentclass[11pt]{article}

\usepackage[final]{acl}

\usepackage{times}
\usepackage{latexsym}
\usepackage{graphicx}

\usepackage[T1]{fontenc}
\usepackage[utf8]{inputenc}

\usepackage{microtype}

\usepackage{inconsolata}

\usepackage{graphicx}
\usepackage{tikz}
\usetikzlibrary{arrows,arrows.meta,shapes,positioning,babel,calc}
\usepackage{amsmath}
\usepackage{amsfonts}
\usepackage{amssymb}
\usepackage{amsthm}
\allowdisplaybreaks
\usepackage{bbm}
\usepackage{subcaption}
\usepackage{enumitem}

\newcommand{\mespace}{\mathcal{M}}
\newcommand{\uspace}{\mathcal{U}}
\newcommand{\wspace}{\mathcal{W}}
\newcommand{\yspace}{\mathcal{Y}}
\newcommand{\lexicon}{\mathcal{L}}
\newcommand{\matS}{\mathbf{S}}
\newcommand{\matL}{\mathbf{L}}
\newcommand{\matP}{\mathbf{P}}
\newcommand{\matC}{\mathbf{C}}

\DeclareMathOperator*{\argmax}{arg\,max}

\title{Collaborative Rational Speech Act: \\ Pragmatic Reasoning for Multi-Turn Dialog}

\author{\textbf{Lautaro Estienne\textsuperscript{1,2, 4, 6}}\quad \textbf{Gabriel Ben Zenou\textsuperscript{1,4}} \quad
\textbf{Nona Naderi\textsuperscript{2, 4, 6}} \\\textbf{Jackie Chi Kit Cheung\textsuperscript{3, 7}} \quad
\textbf{Pablo Piantanida\textsuperscript{1,3, 4, 5, 6}} \vspace{2mm}\\
    \textsuperscript{1}International Laboratory on Learning Systems (ILLS) \\ \textsuperscript{2}Laboratoire Interdisciplinaire des Sciences du Numérique (LISN) \\
    \textsuperscript{3}Mila - Quebec AI Institute  \quad\textsuperscript{4}CNRS \quad\textsuperscript{5}CentraleSupélec \quad\textsuperscript{6}Université Paris-Saclay \\
   \textsuperscript{7}McGill University, Canada CIFAR AI Chair \\
   \small{ \texttt{lautaro.estienne@universite-paris-saclay.fr}}\\
}

\begin{document}
\maketitle
\begin{abstract}
As AI systems take on collaborative roles, they must reason about shared goals and beliefs—not just generate fluent language. The Rational Speech Act (RSA) framework offers a principled approach to pragmatic reasoning, but existing extensions face challenges in scaling to multi-turn, collaborative scenarios. In this paper, we introduce Collaborative Rational Speech Act (CRSA), an information-theoretic (IT) extension of RSA that models multi-turn dialog by optimizing a gain function adapted from rate-distortion theory. This gain is an extension of the gain model that is maximized in the original RSA model but takes into account the scenario in which both agents in a conversation have private information and produce utterances conditioned on the dialog. We demonstrate the effectiveness of CRSA on referential games and template-based doctor-patient dialogs in the medical domain. Empirical results show that CRSA yields more consistent, interpretable, and collaborative behavior than existing baselines, paving the way for more pragmatically competent language agents.
\end{abstract}

\section{Introduction}

Modeling conversations is central to the development of grounded and useful agentic AI systems, which are increasingly characterized by collaborative interactions between humans and machines. Several applications benefit from dialog systems capable of natural interactions with users. For instance, in the medical domain, conversational agents could support diagnostic interviews~\citep{tu_towards_2025} or serve as tools for physician training in controlled environments~\citep{karunanayake_next-generation_2025}. In enterprise settings, dialog agents could autonomously handle routine tasks—such as scheduling, data entry, or report generation—freeing human effort for higher-level decision-making~\citep{tupe_ai_2025,satav_enterprise_2025}. In education, they offer the potential to personalize content delivery, adapting to learners' styles and paces~\citep{nabhani_role_2025,vorobyeva_personalized_2025}. While such applications are still emerging, a key enabler is the development of models that can manage collaborative, goal-oriented interactions in a robust and interpretable manner.
%In enterprise workflows, such agents may autonomously handle routine tasks—like scheduling, data entry, or report generation—freeing up human effort for more strategic decision-making~\citep{tupe_ai_2025,satav_enterprise_2025}. In education, dialog systems could potentially enhance engagement by helping personalize content delivery to better match individual learning styles and paces~\citep{nabhani_role_2025,vorobyeva_personalized_2025}. While many of these applications remain aspirational, developing models that can robustly and interpretably manage collaborative interactions is a crucial step toward their realization.

To succeed in real-world settings, dialog generative-based models must do more than generate fluent language—they must track shared tasks to communicate meaningfully in context~\citep{lin_decision-oriented_2024}. For example, a physician in a diagnostic exchange refines hypotheses as the conversation evolves, requiring interpretable and scalable frameworks for reliable interaction.

%To be effective in real-world settings, dialog models must not only produce fluent responses but also maintain an internal representation of the shared task. This may enable the generation of messages that convey meaning more effectively to the interlocutor~\citep{lin_decision-oriented_2024}. For example, a physician engaged in a diagnostic conversation needs to form a well-grounded estimate of possible pathologies and procedures as the interaction progresses. In tasks like these, a scalable and interpretable framework can support more robust, transparent, and reliable communication. 

Yet, many existing models prioritize task-specific response generation~\citep{he_learning_2017,jiang_towards_2019,fair_human-level_2022}, or optimize for superficial conversation properties using narrowly defined objectives~\citep{khani_planning_2018,dafoe_open_2020,lin_decision-oriented_2024,jeon_reward-rational_2020}. While these methods often yield strong performance, they typically lack principled foundations, leading to task-specific solutions that struggle to generalize or remain robust under shifting conditions.

%Despite the importance of these features, many existing conversational models focus primarily on task-specific language generation~\citep{he_learning_2017,jiang_towards_2019,fair_human-level_2022}, or adopt learning objectives designed to elicit conversations with particular characteristics~\citep{khani_planning_2018,dafoe_open_2020,lin_decision-oriented_2024,jeon_reward-rational_2020}. While these approaches can yield strong empirical results, they often lack theoretical justification and rely heavily on task-specific design choices.

The Rational Speech Act (RSA) framework~\citep{frank_predicting_2012} offers a principled foundation for modeling pragmatic reasoning as recursive social inference between speakers and listeners. Viewed through an information-theoretic (IT) lens, RSA approximates a rate-distortion solution~\cite{DBLP:books/wi/01/CT2001}, where the listener reconstructs intended meaning from observed utterances~\citep{zaslavsky_ratedistortion_2021}. RSA has successfully captured phenomena such as reference~\citep{degen_when_2020}, implicature~\citep{bergen_pragmatic_2016}, and vagueness~\citep{herbstritt_complex_2019}, and powered applications from grounded captioning~\citep{cohn-gordon_pragmatically_2018} to controlled generation~\citep{wang_rsa-control_2024}. Yet, despite this promise, existing RSA extensions remain limited in multi-turn, task-oriented dialog: they struggle to model evolving beliefs or integrate dialog history~\citep{carenini_towards_2024,degen_rational_2023}. We argue this shortfall stems from the absence of a unified, theoretically grounded mechanism for belief and task tracking in collaborative interaction.

%One framework that has been widely used for modeling pragmatic reasoning in collaborative dialogs and is grounded in information theoretic concepts is the Rational Speech Act (RSA) model~\citep{frank_predicting_2012}. RSA formalizes communication as recursive social inference between speakers and listeners, providing a structured way to represent the intended meaning behind an utterance. From an information-theoretic (IT) perspective, the model can be seen as a solution of a Rate-Distortion problem in which the listener attempts to reconstruct this meaning based on the observed message~\citep{zaslavsky_ratedistortion_2021}. The RSA model has been proven to model several linguistics phenomena like reference~\citep{degen_when_2020}, implicature~\citep{bergen_pragmatic_2016} or vagueness~\citep{herbstritt_complex_2019}, and has been applied to a variety of tasks like cross-cultural reference~\citep{shaikh_modeling_2023}, grounded caption generation~\citep{cohn-gordon_pragmatically_2018} and controlled text generation~\citep{wang_rsa-control_2024}, the last two involving multi-step generation. However, traditional RSA models typically lack mechanisms to incorporate dialog history or to track how task-related beliefs evolve over multiple turns~\citep{carenini_towards_2024,degen_rational_2023}. We believe that these limitations may be due in part to the fact that most existing variants rely on heuristics or task-specific designs, lacking a unified and theoretically grounded framework.

%\subsection*{Contributions}
In this paper, we introduce the Collaborative Rational Speech Act (CRSA), an IT grounded extension of the RSA framework designed to model multi-turn, collaborative dialogs. CRSA optimizes a gain function that is a generalization of the one proposed by  \citet{zaslavsky_ratedistortion_2021} for the multiple-turns scenario. The resulting model provides a tool to model an estimation of the target of the joint task and the belief that each agent has on their in interlocutor's private information, and it can be used with large language models (LLMs). We evaluate CRSA in referential game settings and semi-automatic generated conversations between doctor and patients for extracting a medical diagnosis.\footnote{Code is available at \texttt{\url{https://github.com/LautaroEst/crsa}}}

Our main contributions are as follows:
\begin{itemize}[noitemsep, topsep=0pt, leftmargin=*]
\item We introduce \textbf{Collaborative RSA (CRSA)}, a novel, information-theoretically grounded extension of the RSA framework tailored for multi-turn, goal-driven dialog. 
    
\item \textbf{A generalized multi-turn gain function:} We extend the rate-distortion to model multi-turn collaborative settings of RSA, capturing both task progression and evolving partner beliefs. CRSA jointly models the agent’s belief about (i) the shared task target and (ii) the interlocutor’s private knowledge—enabling socially aware and context-sensitive communication.

%\item We \textbf{demonstrate CRSA’s effectiveness} on referential games and doctor--patient dialogs, showing improved consistency and collaboration. We also show how CRSA integrates with LLMs to guide generation through belief-aware objectives.
 
\item \textbf{Empirical validation:} We evaluate CRSA on referential games and semi-automatically generated doctor-patient dialogs, showing that it improves consistency, interpretability, and collaborative alignment compared to existing baselines.  %We illustrate how CRSA can be integrated with LLMs, steering their outputs via belief-guided gain optimization.
  
\end{itemize}

\section{Related work}

\paragraph{RSA model and pragmatics.}
The Rational Speech Act (RSA) framework~\citep{frank_predicting_2012} serves as a model for pragmatic communication designed to emulate human behavior in linguistic tasks~\cite{degen_when_2020,bergen_pragmatic_2016,herbstritt_complex_2019,spinoso-di-piano-etal-2025-rsa}. 
This framework is both conceptually intuitive and computationally versatile, making it readily adaptable for integration with neural language models to tackle more intricate challenges, including machine translation~\citep{cohn-gordon_lost_2019}, image captioning~\citep{cohn-gordon_pragmatically_2018}, controllable text generation~\citep{shen_pragmatically_2019,wang_rsa-control_2024,darrin-etal-2024-glimpse}. Extensions to the original RSA framework have been proposed to accommodate more complex scenarios. For instance, adaptations have addressed cases where agents lack shared vocabularies~\citep{bergen_pragmatic_2016} or where common ground evolves dynamically during interaction~\citep{degen_wonky_2015}.
A comprehensive overview of RSA's development and its numerous variants is provided by \citet{degen_rational_2023}.

\paragraph{Information-theoretic results for interactive rate-distortion.}
Information theory offers a robust framework for analyzing communication as the exchange of information between agents. Within this domain, the rate-distortion problem~\citep{5311476} offers a principled way to balance compression efficiency with the fidelity of reconstruction. This problem has been pivotal in exploring the trade-offs between fidelity and compression in message transmission. \citet{kaspi_two-way_1985} investigated scenarios involving two agents engaging in iterative interactions to collaboratively infer each other's observations. Building on this foundation, \citet{rey_vega_three-terminal_2017} extended the analysis to multi-agent contexts, accommodating communication frameworks with three or more participants and significantly advancing the understanding of collective information exchange.
Focusing on two-agent systems, \citet{vera_collaborative_2019} explored a variation wherein each agent is tasked not merely with understanding one another but with predicting a target random variable representing a (possible stochastic) function of  each other's observations. This approach highlights the promise of IT methods in supporting more efficient and collaborative communication among agents in complex environments, as shown by \citet{zaslavsky_ratedistortion_2021}, who reformulate the standard RSA framework as a rate-distortion optimization problem.

\paragraph{Collaborative dialog modeling.}
Multiple works frame a collaborative or task-oriented dialog as a Partially Observable Markov Decision Process (POMDP)~\citep{williams_partially_2007}, which provide a suitable framework to model end-to-end networks on specific tasks~\citep{wen_network-based_2017,jiang_towards_2019}. Reinforcement learning has been widely used in this context in order to provide interpretable and trackable training procedures that incorporate the structure of the dialog in their policy training or decoding strategy~\citep{lin_decision-oriented_2024,li_deep_2016,xu_efficient_2025}. Related to this, game-theoretic perspective has also been used in dialog modeling~\citep{jeon_reward-rational_2020,lin_inferring_2022}. In this context, multiple tasks and datasets have been developed to evaluate dialog modeling~\citep{he_learning_2017,khani_planning_2018,macherla_mddial_2023}, usually by assessing the task performance and the similarity with human conversations. The RSA model has also found applications in dialog systems, often complementing neural models to enhance agent consistency given persona~\citep{kim_will_2020} or to improve the interpretation of emotional subtext~\citep{kim_perspective-taking_2021}.

\section{Review of the RSA Model from the Lens of Information Theory}

\begin{figure}[t]
    \centering

    \begin{subfigure}[t]{0.35\textwidth} % width of each subfigure
        \centering
        \includegraphics[width=\textwidth]{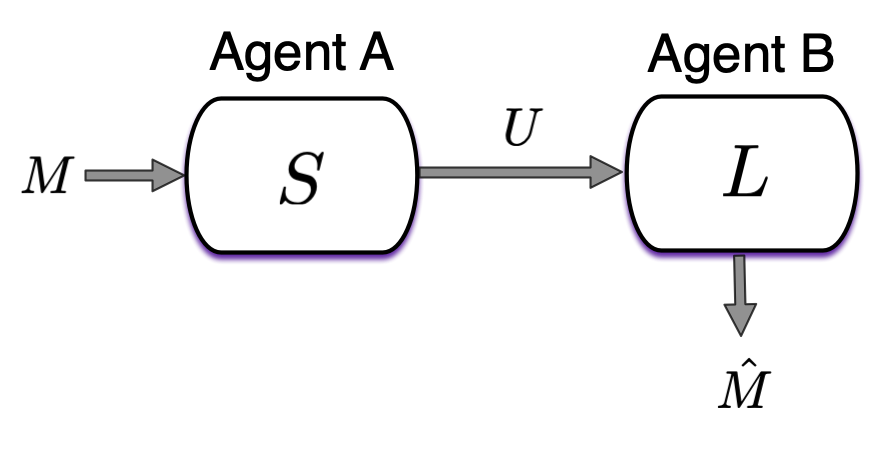}
        \caption{RSA}
        \label{fig:rsa_diagram}
    \end{subfigure}

    \vspace{1em}

    \begin{subfigure}[t]{0.45\textwidth}
        \centering
        \includegraphics[width=\textwidth]{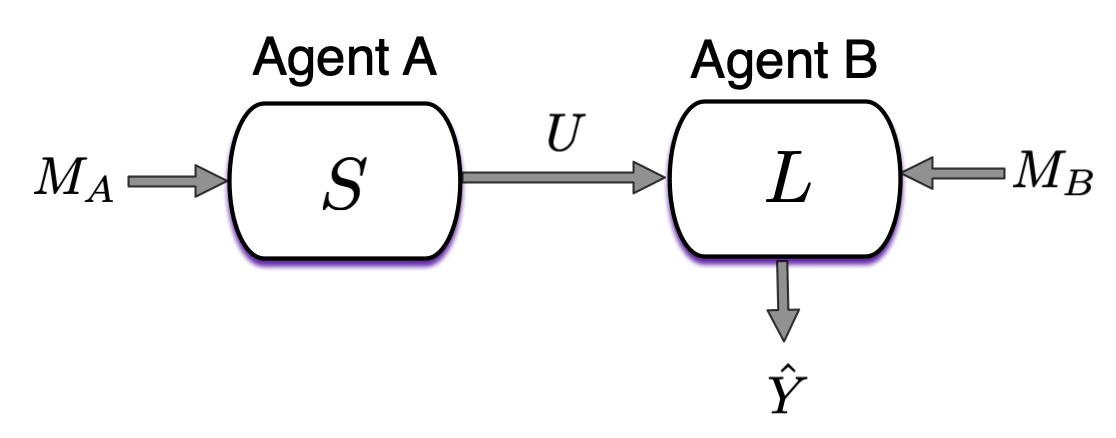}
        \caption{YRSA}
        \label{fig:yrsa_diagram}
    \end{subfigure}

    \vspace{1em}

    \begin{subfigure}[t]{0.5\textwidth}
        \centering
        \includegraphics[width=\textwidth]{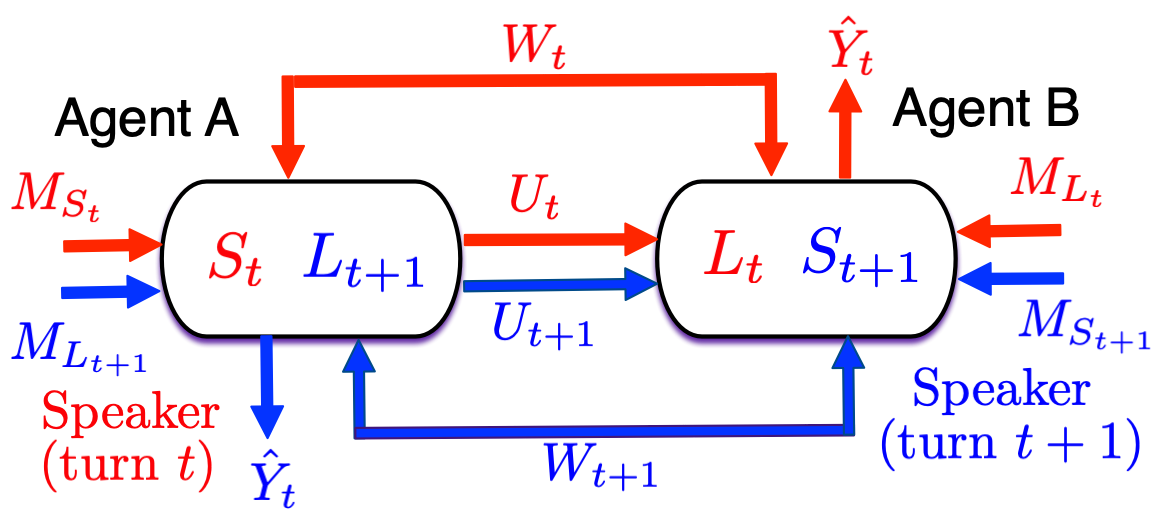}
        \caption{CRSA}
        \label{fig:crsa_diagram}
    \end{subfigure}

    \caption{RSA variants proposed in this work (YRSA \subref{fig:yrsa_diagram}, CRSA \subref{fig:crsa_diagram}) compared to the original one (RSA \subref{fig:rsa_diagram}).}
    \label{fig:models_diagrams}
\end{figure}

% TODO: CHANGE Y --> Y_t

Figure~\ref{fig:rsa_diagram} presents a schematic view of the classic RSA model from an information-theoretic perspective. Here, a meaning $m\in\mespace$ is received by the speaker $S : \mespace \times \uspace \rightarrow [0,1]$ who uses it to produce a posterior probability $S(u|m)$ for all possible utterances $u\in\uspace$. The utterance $u$ is then transmitted to the listener $L : \uspace \times \mespace \rightarrow [0,1]$ who produces a posterior $L( m|u)$ for all possible reconstructions $ m\in\mespace$ of the meaning $m$ that the speaker is trying to convey. Additionally, there is a distribution $P:\mespace \rightarrow [0,1]$ that is known by the two agents and represents the prior of the meanings. Finally, the function $C : \uspace \rightarrow \mathbb{R}$ assigns a prior cost value to each utterance produced by the speaker.\footnote{Note that the prior and cost functions are not shown in Figure~\ref{fig:models_diagrams} for clarity.}

In the classic RSA model, agents update their values based on the other’s perspective. For simplicity, and without loss of generality, we adopt the listener's viewpoint—assuming the speaker updates first\footnote{In the classic RSA literature, the literal listener (speaker) is usually represented with $L_0$ ($S_0$) and the pragmatic with $L_1$ ($S_1$). Here, we will reserve the subindex notation for the turn number and denote the level of pragmatism of each agent by using the super index $L^k$ ($S^k)$ with $k=0,1,\ldots,K$.}:
\begin{align*}
    S^{k+1}(u| m) &\propto \exp \big[\alpha (\log L^k( m|u)- C(u))\big], \\
    L^{k+1}( m|u)&\propto S^{k+1}(u| m)P( m).
\end{align*}
In this case, the listener is initialized with a predefined lexicon function $\lexicon : \uspace \times \mespace \rightarrow \{0,1\}$, which specifies the possible meanings associated with each utterance:
\begin{equation*}
    L^0( m|u) \propto P( m)\lexicon(u, m). 
\end{equation*}
\citet{zaslavsky_ratedistortion_2021} show that this iteration process is equivalent to maximizing the following objective: 
\begin{align}\label{eq:rsa_gain}
    \mathcal{G}_{RSA}^{\alpha}(L,\!S)\!=\!H_S(U| M)\!+\!\alpha \mathbb{E}_S[V_L(U,\! M)],
\end{align}
where $H_S(U| M)$ is the conditional entropy between the estimated meanings and the utterances, $V_L(u, m)\triangleq \log L( m|u)\!-\!C(u)$ is called the ``listener value'', and $\mathbb{E}_S[V_L]$ is computed with respect to the distribution of the speaker. That is,
\begin{align*}
    H_S(U| M)& \! =\! -\!\sum_{\forall\,(u, m)}P_S(u, m)\!\log\!S(u| m), \\
    \mathbb{E}_S[V_L]& \!=\!\!\sum_{\forall\,(u, m)}\!P_S(u, m) V_L(u, m),
\end{align*}
where $P_S(u, m)\triangleq S(u| m)P( m)$ represents the joint probability of the speaker.

\section{Main Theoretical Results}
\subsection{Modeling private meanings (YRSA)}\label{sec:yrsa_model}

To extend the RSA model to bidirectional dialog with explicit task modeling, we first distinguish between private meanings and shared task outcomes. In real conversations, each participant holds their own prior knowledge and worldview, which may differ from that of their interlocutor. In our example of a dialog between a patient and a physician: the patient must describe their symptoms, which are not directly observable by the physician, while the physician brings medical expertise the patient lacks. Both types of knowledge are essential for determining the appropriate diagnosis or treatment plan. Notably, neither the patient’s symptoms nor the physician’s prior knowledge fully capture the shared goal of the conversation, i.e. the identification of a suitable medical outcome.

In this context, we identify the need of representing a private set of meanings $\mespace_A$ and $\mespace_B$ for each agent, which may or may not match. In addition, the result $y$ of the shared task is going to be represented with a separate space $\yspace$ that contains all the possible outcomes of it. For simplicity, we will assume that all these are discrete spaces. Figure~\ref{fig:yrsa_diagram} represents a schematic of this model. We will refer to this extension as the YRSA model. 

The YRSA model redefines the notion of prior from the classic RSA framework by conditioning the dialog on the joint realization of the agents’ private meanings ($m_A$, $m_B$) and the shared task target $y$, which together define the context in which the interaction unfolds. Importantly, we assume for the development of our  model that both the realizations and the joint distribution of these three variables do not change over time during the conversation.  This implies that the prior is completely defined by the joint distribution $P : \mespace_A \times \mespace_B \times \yspace \rightarrow [0,1]$ given to both  agents.

We now turn to defining the updated agent posteriors.  The new speaker $S : \mespace_{A} \times \uspace \rightarrow [0,1]$ produces a posterior $S(u|m_A)$ that only depends on its the private meaning $m_A$. Similarly, the listener $L:\mespace_B\times\uspace\times\yspace\rightarrow[0,1]$ is represented by the posterior $L(y|m_B,u)$, which is conditional independent of the private meanings $m_A$. In this formulation, the representation of task performance is delegated to the listener, who updates their belief upon receiving the utterance.

We can now propose the corresponding gain function to be maximized by this model:
\begin{align}\label{eq:yrsa_gain}
    \mathcal{G}_{YRSA}^\alpha(L,S) &= H_S(U|M_A) \nonumber \\
    &+ \alpha \mathbb{E}_S[V_L(U,M_B,Y)]
\end{align}
with $V_L(u,m_B,y) = \log L(y|u,m_B) - C(u)$ and $H_S(U|M_A)$ defined as in the classic RSA. A detailed derivation of the equations used to maximize this function is provided in Appendix~\ref{app:yrsa}.

\subsection{The CRSA Model}\label{sec:crsa}

Effective collaboration requires not only modeling agents’ private meanings and the shared task, but also supporting multi-turn dialog. In a medical consultation, for instance, the patient shares symptoms and background, while the physician asks questions, proposes diagnoses, and recommends treatments. To capture such interactions, we denote the speaker’s utterance at turn $t$ as $U_t$, and the dialog history up to that point as $W_t = (U_1, \ldots, U_{t-1})$, representing the sequence of prior exchanges.

%Back to the case study of a medical consultation, not only each agent has its own expertise or private knowledge, but it also alternately shares information to the other. The patient details its symptoms, the current medicines it's taking, its past days and the possible sources of its sickness, while the physician asks specific questions, names possible diseases, and suggests potentially effective medication to treat it.

%To model the several turns making up the dialog, we define the conversation history denoted by $W_t=(U_1,\ldots,U_{t-1})$ as all the past utterances up to the current turn $t$. 

%\subsubsection{Modeling multi-turn dialog with explicit history (baseline)}
The attempt of previous approaches to incorporate the history of the conversation to the RSA model relies on defining the lexicon (or directly the literal listener/speaker) as a function of each turn~\citep{wang_rsa-control_2024,kim_will_2020,lin_inferring_2022}. In many cases, this lexicon is given by the output of a neural language model and can be very robust to the evolving dialog. However, that variant of the RSA does not correspond to maximizing the gain of Equation~\eqref{eq:rsa_gain}, but a modified version of it in which $U_t$ is replaced by $(U_t,W_t)$:
\begin{align}\label{eq:rsa_naive}
    H_S(U_t,\!W_t| M)\! +\! \alpha \mathbb{E}_S[V_L(U_t,\!W_t,\! M,\!Y)].
\end{align}
This is equivalent to applying an RSA model at each turn by initializing it with a lexicon $\lexicon(u_t, m, w_t)$ depending on $w_t$, the past utterances. 

The issue with Equation~\eqref{eq:rsa_naive} is that the speaker's utterance $U_t$ at turn $t$ is modeled \emph{jointly} with the dialog history $W_t$, rather than being explicitly \emph{conditioned} on it. To express the gain in terms of the conditional entropy of the current utterance alone, we condition it on both the dialog history $W_t$ and the speaker's intended meaning $M$, rather than on $M$ alone. In Section~\ref{sec:crsa_eq}, we formally introduce the corresponding expressions of the CRSA model, which incorporates this notion of multi-turn conditioned to the past utterances, as well as private meanings and target task.

%In order to keep a version of the gain with a conditional entropy of the current utterance only, we decide to model this utterance conditioned on both the past $W_t$ and the meaning of the speaking agent $M_{S_t}$ rather than on the meaning $M_{S_t}$ only. 

\subsubsection{Equations of the CRSA model}\label{sec:crsa_eq} Figure~\ref{fig:crsa_diagram} illustrates our extension of the YRSA model to the collaborative setting. As in the original setup, agents alternate roles—one acting as the speaker, the other as the listener—to achieve a shared task. Each agent has access to a private meaning space, $\mespace_A$ or $\mespace_B$, which remains hidden from their counterpart. Then, at turn $t$, the private meanings of the speaker will correspond to the meanings of the agent playing the role of the speaker and vice-versa. We refer as $\mespace_{S_t}$ and $\mespace_{L_t}$ to the private meanings of the speaker and the listener at turn $t$, respectively. For instance, if speaker A starts the conversation, $\mespace_{S_1}=\mespace_A$ and $\mespace_{L_1}=\mespace_B$. Both agents also have access to the conversation history, denoted as $w_t = (u_1, \ldots, u_{t-1}) \in \wspace_t \triangleq \uspace_1 \times \cdots \times \uspace_{t-1}$, where each $\uspace_i$ represents the space of possible utterances at turn $i$. The shared objective is to jointly predict a target class $y$ from a finite discrete set $\yspace$.

As discussed earlier, the joint distribution $P(m_A, m_B, y)$ serves as a fixed prior throughout the conversation. To maintain consistency as agents alternate roles, we define the prior at turn $t$ over the active speaker and listener meanings, i.e., $P(m_{S_t}, m_{L_t}, y)$, as follows:
\begin{align*}
    P_t(m_{S_t},m_{L_t},y)\!=\!\begin{cases}
        P(m_{S_t},m_{L_t},y) & \!\!\mbox{if } S_t\!=\!A \\
        P^\top(m_{L_t},m_{S_t},y) & \!\!\mbox{if } S_t \!=\!B
    \end{cases}
\end{align*}
where $P^\top: \mespace_B \times \mespace_A \times \yspace \rightarrow [0,1]$ is such as $P^\top(b,a,y)=P(a,b,y)$. This definition simply represents swapping the arguments corresponding to agent A and B to reflect the role change.

 Formally, we define the distribution of each agent at turn $t$. 
 The speaker $S_t : \mespace_{S_t} \times \uspace_t \times \wspace_t \rightarrow [0,1]$ produces a posterior $S_t(u_t|m_{S_t},w_t)$ that depends on its private meaning $m_{S_t}$ and the past utterances $w_t$. On the other hand, the listener $L_t:\mespace_{L_t}\times\uspace_t\times\wspace_t\times\yspace\rightarrow[0,1]$ is represented by the posterior $L_t(y|m_{L_t},u_t,w_t)$ which is independent of the private meanings of the speaker.
 %Formally, we are assuming the following markov chains:
% \begin{align}
%      M_{S_t} \markovchain (U_t,W_t,M_{L_t}) \markovchain Y \\
%      (M_{L_t},Y) \markovchain (M_{S_t},W_t) \markovchain U_t
% \end{align}

Building on the gain function in Equation~\eqref{eq:rsa_gain}, we extend the joint speaker distribution and listener utility to incorporate private meanings and multi-turn dialog:
\begin{align*}
  &  P_S(u_t,w_t,m_{S_t},m_{L_t},y) \triangleq S_t(u_t|m_{S_t},w_t) \times \nonumber \\
    &  P_S(w_t|m_{S_t},m_{L_t})  P_t(m_{S_t},m_{L_t},y), \\[.5em]
   & V_L(u_t,w_t,m_{L_t},y)\!\triangleq\! 
    \log L_t(y | u_t,\! m_{L_t},\! w_t)\!-\!C(u_t).
\end{align*}
% the conditional entropy and the listener value for the new scenario:
% \begin{align}
%     &H_S(U_t|M_S,W_t) = \nonumber \\ &\!\!\sum_{\substack{w_t u_t y\\m_S m_L}}\!\!\!P_S(u_t,\!w_t,\!m_S,\!m_L,\!y)\!\log\! S(u_t|m_S,\!w_t) \\[.5em]
%     &\mathbb{E}_S[V_L] = \nonumber \\
%     &\!\!\sum_{\substack{w_t u_t y\\m_S m_L}}\!\!\!\!P_S\!(\!u_t,\!w_t,\!m_S,\!m_L,\!y\!) V_L(\!u_t,\!w_t,\!m_S,\!m_L,\!y\!)
% \end{align}
Then, we define one gain function at each turn to be maximized:
\begin{align}
    \mathcal{G}_{CRSA}^{\alpha}&(L_t,S_t) =  H_{S_t}(U_t|M_{S_t},W_t) \nonumber \\ &+ \alpha \mathbb{E}_{S_t}[V_L(U_t,\!W_t,\!M_{S_t},\!M_{L_t},\!Y)],
\end{align}
where the expectation of both terms is over $P_S$. In all cases,  we will  model $P_S(w_t|m_{S_t},m_{L_t})$ with the past speakers' utterances:
\begin{align}\label{eq:belief_decomposition}
    &P_S(w_t|m_{S_t},m_{L_t}) = \nonumber \\ &\underbrace{\prod_{\substack{i<t\\S_i=S_t}}S_i(u_i|w_i,m_{S_t})}_{B_{L,t}(m_{S_t})} \underbrace{\prod_{\substack{i<t\\S_i\neq S_t}}S_i(u_i|w_i,m_{L_t})}_{B_{S,t}(m_{L_t})}.
\end{align}
This formulation naturally leads to interpreting $B_{L,t}(m_{S_t})$ and $B_{S,t}(m_{L_t})$ as each agent’s belief about their interlocutor’s private meaning. In Section~\ref{sec:find_a1}, we illustrate why this interpretation is reasonable with a concrete example.

Once modeled the gain, the equations that correspond to its maximization are the following:
% \begin{align}
%     & S^{k+1}_{t}(u_t|w_t,m_{S_t}) \propto  \exp(\alpha\sum_{m_{L_t},y}\frac{P_S(w_t|m_{S_t},m_{L_t})}{P_S(w_t|m_S)}P_t(y,m_{L_t}|m_{S_t})\log L^{k}_t(y|u_t,w_t,m_{L_t}) - C(u_t)) \\[.5em]
%     & L^{k+1}_t(y|u_t,w_t,m_{L_t}) \propto \sum_{m_{S_t}}P_S(w_t|m_S,m_L)P_t(m_{S_t},m_{L_t},y) S^{k+1}_t(u_t|w_t,m_{S_t})
% \end{align}
\begin{align*}
    & S^{k+1}_{t}(u_t|w_t,m_{S_t}) \propto  \nonumber \\ &\exp\big[\alpha\!\!\sum_{\forall (m_{L_t},y)}\!\!\!\!B'_t(m_{S_t},m_{L_t},y)V_L(u_t,w_t,m_{L_t},y) \big],\\[.5em]
    & L^{k+1}_t(y|u_t,w_t,m_{L_t}) \propto \nonumber \\
    &\sum_{\forall m_{S_t}}B_{L,t}(m_{S_t})P_t(m_{S_t},m_{L_t},y) S^{k+1}_t(u_t|w_t,m_{S_t}),
\end{align*}
where we replace\footnote{For simplify notation, we removed the $t$ subindex.} $ B'_t(m_S,m_L,y) =$
\begin{equation}\label{eq:belief_prime}
      \frac{B_{S,t}(m_L)P(m_L|m_S)}{\sum_{\forall m_L'}B_{S,t}(m_L')P(m_L'|m_S)}P(y|m_L,m_S). 
\end{equation}
A complete derivation of these equations is provided in Appendix~\ref{app:proof}. Finally, there is no single prescribed method for initializing the iteration at each turn. In Section~\ref{sec:find_a1}, we adopt the listener's perspective and explore two variants of the initial lexicon $\lexicon$, initializing the literal listener as: 
\begin{align}\label{eq:literal_listener}
    L^0(y|u_t,&w_t,m_{L_t}) \propto \nonumber \\ &\sum_{\forall\, m_S}P(m_S,m_{L_t},y)\lexicon_{u_t,w_t}(m_{S_t})
\end{align}
with $\lexicon_{u_t,w_t}(m_{S_t})$ depending on the variant of the RSA. In contrast, in Section~\ref{sec:mddial} we initialize the literal speaker directly with a LLM:
\begin{equation}\label{eq:literal_speaker}
    S_t^0(u_t|m_{S_t},w_t) \propto P_{LM}(u_t|w_t,\mathrm{prompt}(m_{S_t})),
\end{equation}
where $\mathrm{prompt}(m_{S_t})$ is the text used to prompt the speaker at that turn. As shown, CRSA retains the flexibility of the original RSA framework in modeling both the listener's and speaker's perspectives.

\paragraph{Algorithmic complexity.} We find that at turn $t$, these new set of equations scale as $\mathcal{O}\left(K \cdot |\mespace_A| \cdot |\mespace_B| \cdot |\yspace| \cdot |\uspace_t|\right)$ where $K$ is the number of iterations to produce the pragmatic agents. In contrast, the classic RSA equations scale as $\mathcal{O}\left(K \cdot |\mespace| \cdot |\uspace|\right)$.

\section{CRSA for Reference Games}\label{sec:find_a1}

To evaluate CRSA, we adapt the reference game of \citet{khani_planning_2018}. In this setting, two agents are shown the same sequence of $N$ cards, each labeled with one letter (A or B) and one number (1 or 2). Agent A sees only the letter on each card, while Agent B sees only the number. Their goal is to collaboratively identify the position of the card labeled A1. At each turn, an agent may utter a number from $1$ to $N$, indicating a card position. For simplicity, we assume that each round contains at most one A1 card and that Agent A always initiates the exchange.

%For our first assessment of the CRSA model we designed a reference game by using the game introduced in \citet{khani_planning_2018}. In this game, two agents are given the same set of $N$ cards, each of which contains exactly one letter (A or B) and one number (1 or 2). However, the first agent (agent A) can only see the letter in the card and the second agent (agent B), the number of the card. The cards are put in a sequence and both agents have to communicate to find out the position of the card with value A1. At each turn, one agent can only say a number from $1$ to $N$ representing the position of a card. For simplicity, we will assume that in each turn there is always exactly one or zero cards with the value $A1$ and that agent A always start the round.

\subsection{Experimental set-up}

For this simulation, we consider that the set $\uspace_t$ of possible utterances at turn $t$ is always ($\forall t$) the set $\uspace_t=\{1,\ldots,N\}$ representing the messages of the form \emph{``A1 card may be at position $n$''} with $n\in\uspace_t$. For the set $\yspace$ of possible classes, the results can be as well \emph{``A1 card may be at position $n$''}, with the addition that there is also the possibility of \emph{``There is no A1 card''}. That is, $\yspace=\{0,1,\ldots,N$\} with $0$ representing the mentioned possibility. 
Regarding the meaning spaces, they correspond to the possible sequences of length $N$ that can be obtained combining without replacement the letters A and B (for agent A) and the numbers 1 and 2 (for agent B). That is, for instance if $N=3$, $\mespace_A=\{\textrm{AAA},\textrm{AAB},\ldots,\textrm{BBB}\}$ and $\mespace_B=\{111,112,\ldots,222\}$. Finally, the prior distribution $P(m_A,m_B,y)$ can be defined as follows:
\begin{align*}
    P(m_A,m_B,y)\propto \begin{cases}
        1 &\mbox{if $m_A$ and $m_B$ form} \\[-.3em]
        &\mbox{A1 at position $y$} \\
        0 &\mbox{otherwise}
    \end{cases}
\end{align*}
Since this is a reference game, we adopt the listener's perspective. In all cases, the literal listener is initialized using Equation~\eqref{eq:literal_listener}, and different model variants are defined based on the update equations and the specification of the lexicon $\lexicon_{u_t,w_t}(m_{S_t})$.

\begin{itemize}[noitemsep, topsep=0pt, leftmargin=*]
    \item \textbf{CRSA}: We apply the CRSA update equations and
    define a lexicon $\lexicon_{u_t,w_t}(m_{S_t})= \lexicon(u_t,m_{S_t})$ that do not depend on $w_t$:
    \begin{align}\label{eq:lexicon_not_wt}
        \lexicon(u_t,m_{S_t}) = \begin{cases}
            1 & \text{if $m_{S_t}$ contains A (or 1) at} \\[-.5em]
            & \text{position $n$ and $u_t=n$}\\
            1 & \text{if there is no A (or 1) in $m_{S_t}$}\\
            0 & \mbox{otherwise}
    \end{cases}
    \end{align}
    \item \textbf{CRSA-$W_t$}: We apply the CRSA update equations, but with the lexicon $\lexicon_{u_t,w_t}(m_{S_t})= \lexicon(u_t,m_{S_t},w_t)$ depending on the the past $w_t$. To define $\lexicon(u_t,m_{S_t},w_t)$, we follow the simple rule:
    \begin{align}\label{eq:lexicon_wt}
        \lexicon(u_t,m_{S_t},w_t) = \begin{cases}
            0 &\mbox{if $u_t \in w_{t-1}$} \\[-.3em]
            &\wedge \; u_t \neq u_{t-1} \\
            \lexicon(u_t,m_{S_t}) &\mbox{otherwise}
        \end{cases}
    \end{align}
We expect efficient conversational behavior in this game to involve repeating an utterance only to confirm the correct A1 card position. If the correct position is identified, agents should repeat the utterance until the round ends; otherwise, repeating it would be inefficient. The rule in Equation~\eqref{eq:lexicon_wt} explicitly encodes this behavior. 
    \item \textbf{YRSA}: We initialize the listener using the YRSA iterative equations and the lexicon from Equation~\eqref{eq:lexicon_not_wt}, effectively applying the RSA iteration in the setting where each agent holds a private meaning—that is, the standard YRSA setup.
    \item \textbf{YRSA-$W_t$}: The same as the one above but using Equation~\eqref{eq:lexicon_wt} as lexicon instead of Equation~\eqref{eq:lexicon_not_wt}.
    \item \textbf{Literal}: In this case, there is no iteration and we simply use Equation~\eqref{eq:literal_listener} to predict the target. We use lexicon of Equation~\eqref{eq:lexicon_not_wt}.
    \item \textbf{Literal-$W_t$}: This is the same as above but using Equation~\eqref{eq:lexicon_wt} as lexicon. 
    \item \textbf{Prior}: In this case, we compute $P(y|m_{L_t})$ from $P(m_{S_t},\!m_{L_t},y)$ for all turns instead of $L_t(y|u_t,m_{L_t},w_t)$. %\textcolor{red}{\# Why P  does not have t here ? \#} 
    This case does not account for the dialog or the current utterance. %\textcolor{red}{\# this sentence is not clear. Are you doing argmax ? Are expecting this distribution to be degenrated ? How are you decoding ? \#}
\end{itemize}

\subsection{Numerical results and discussion}

\begin{figure}[t]
    \centering
    \includegraphics[width=.4\textwidth]{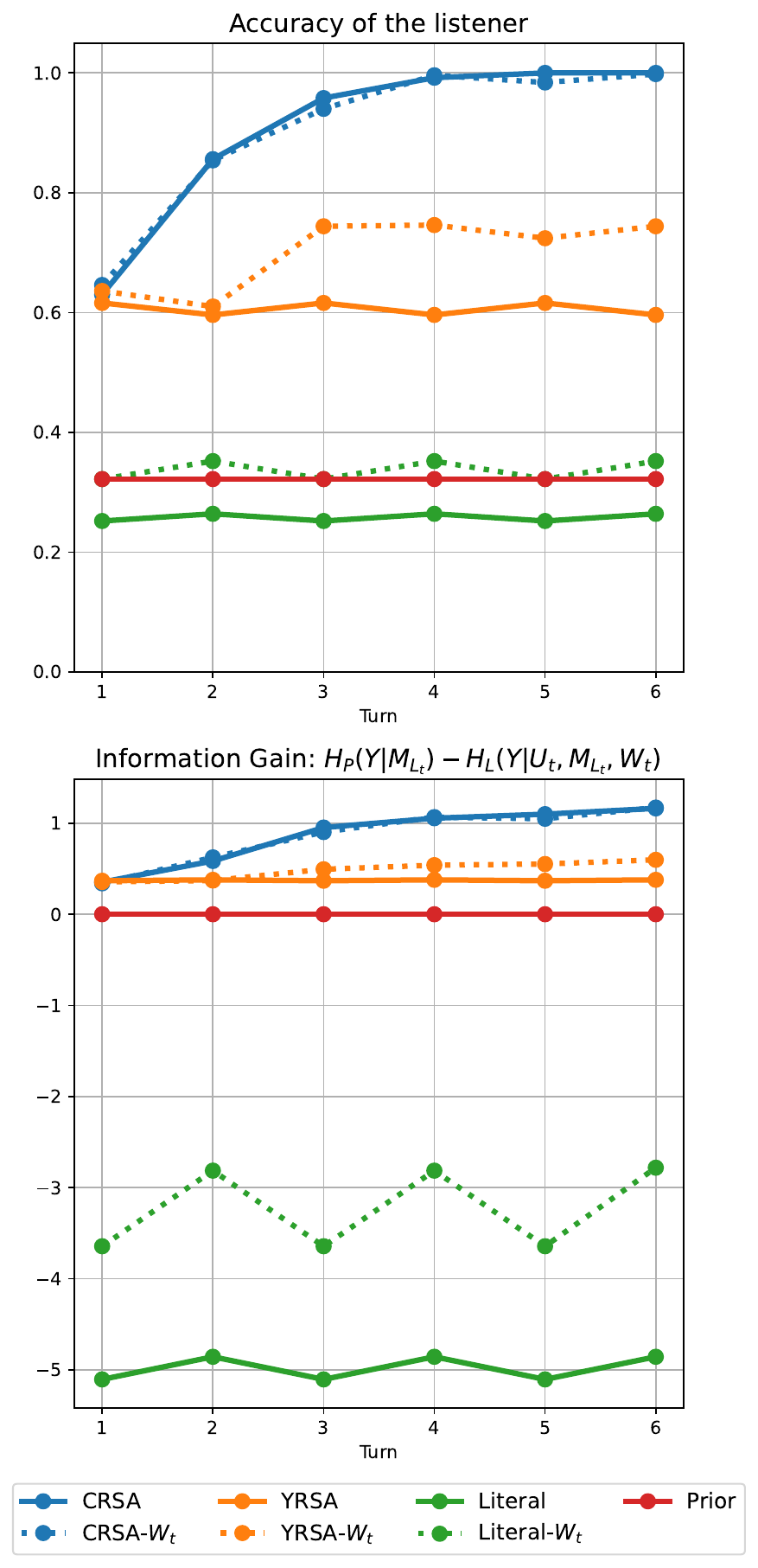}
    \caption{Average of correct predictions with the listener value (top) and information gain (bottom) for 500 rounds of the reference game.}
    \label{fig:find_a1_performance}
\end{figure}

\begin{figure}[t]
    \centering
    \includegraphics[width=.5\textwidth]{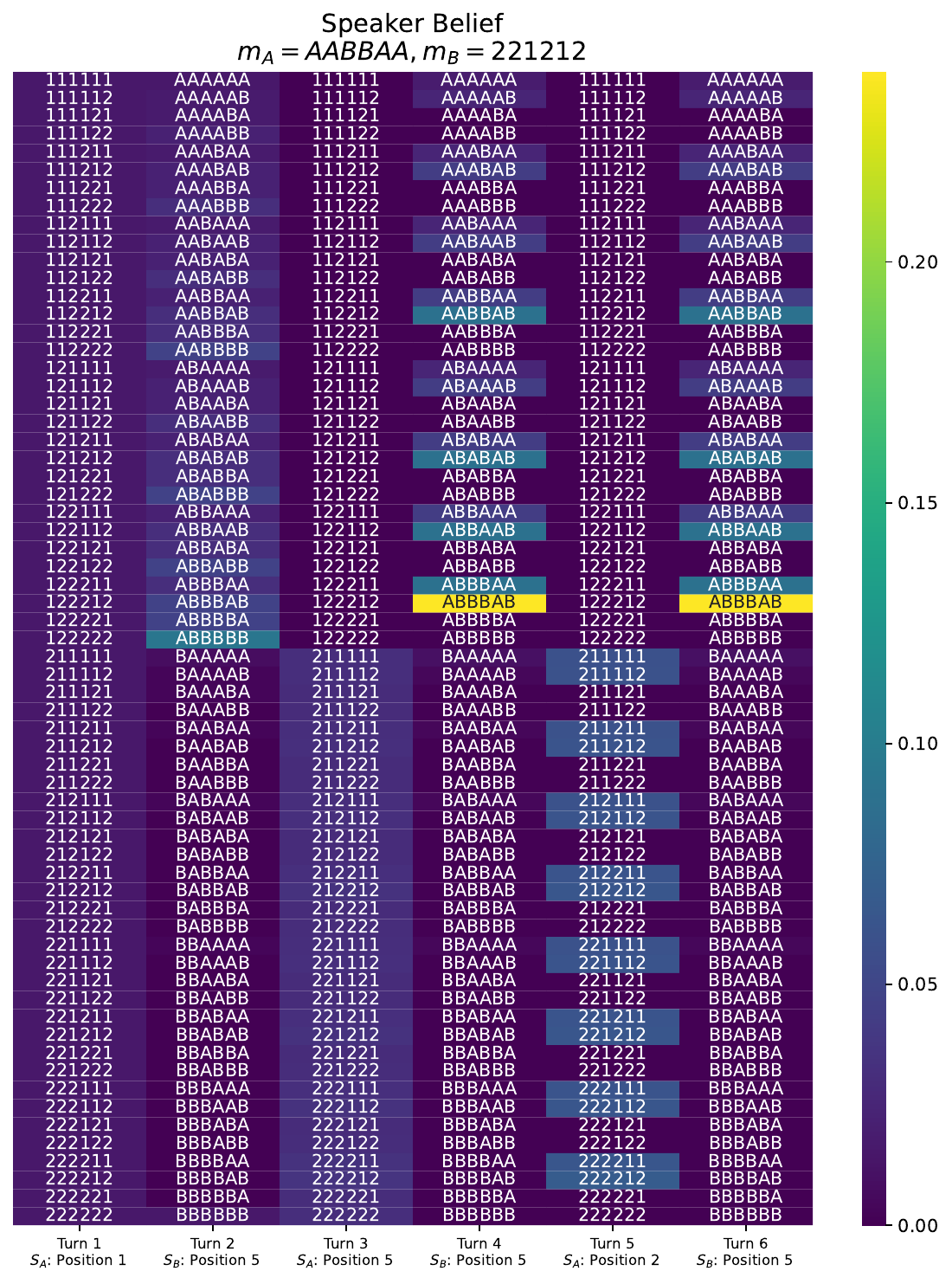}
    \caption{Internal belief of both agents.}
    \label{fig:find_a1_belief}
\end{figure}

Figure~\ref{fig:find_a1_performance} presents the performance of the CRSA model compared to baseline models for $\alpha = 2.5$. Each curve corresponds to a different model evaluated over 500 rounds of the game. The top plot displays task accuracy, measured as the proportion of correct guesses obtained by taking the argmax of the listener’s posterior probability. As accuracy may not be fully representative of the confidence on the decision made by the listener, we also show the \emph{Information Gain} (in the bottom plot) for each turn $t$, computed as the difference $\textrm{IG}(L_t) = H_P(Y|M_{L_t})-H_L(Y|U_t,M_{L_t},W_t)$. That is, given a set of $N$ rounds (all with the same number of turns), the listener's conditional entropy is defined as $H_L(Y|U_t,M_{L_t},W_t)=-1/N\sum_{i=1}^N\log L_t(y^{(i)}|u^{(i)}_t,w^{(i)}_t,m^{(i)}_{L_t})$, and the conditional entropy of the prior is defined as $H_P(Y|M_{L_t})=-1/N\sum_{i=1}^N\log P(y^{(i)}|m^{(i)}_{L_t})$, where the super-index $(i)$ denotes the value at round $i$. As $P(y|m_{L_t})$ takes no account for the interchanged utterances, this metric could be interpreted as the amount of information gained by using the utterances of the dialog up to turn $t$. For all models where there is iteration, we run the model until the gain converged using a tolerance of $1e-3$, so the number of iterations may vary between each turn. We tried various values of $\alpha > 1$ and all values showed best performance of the CRSA model. For values $\alpha\leq1$, all iterative algorithms always produced uniform distributions.

As shown in the plots, the CRSA model outperforms all baselines across both metrics. Moreover, incorporating a lexicon that depends on the past $w_t$ neither improves nor diminishes performance, suggesting that the information encoded in Equation~\eqref{eq:lexicon_wt} is already effectively captured by the CRSA model.  In contrast, the information in Equation~\eqref{eq:lexicon_wt} is not captured by the YRSA-$W_t$ model, which appears to improve as the conversation progresses. As expected, models that do not incorporate dialog history maintain consistent performance across turns, with variations driven only by role changes. We also observed that the CRSA model's variance decreases over time, although this is not shown in the plots for clarity.

Figure~\ref{fig:find_a1_belief} presents an example of a dialog between the agents, along with their internal belief states at each turn. This dialog was generated by sampling the pragmatic speaker distribution at each turn. Each column displays the value of $B_{S,t}(m_{L_t})$ for each possible meaning $m_{L_t}$ of the listener at turn $t$. Notably, as the conversation progresses, the meanings associated with previously uttered messages tend to gain higher belief values, reflecting a refinement in the speaker's inference about the listener's state. We note that speaker at turn 5 produce the "Position 2" utterance, which is a little uninintuitive. However, since utterances are sampled from the full pragmatic speaker distribution rather than chosen greedily and "Position 2" has non-zero probability, this utterance was drawn randomly out of other possible utterances. Such decoding strategy can be interpreted as an exploratory strategy. Importantly, Agent B's return to "Position 5" in the following turn is consistent with its high posterior belief. We also note that the value that maximizes  $B_{S,t}(m_{L_t})$  at turn 6 does not correspond exactly to the correct meaning, but it is a close approximation since the utterance ``Position 6'' never occurred during the round. This supports interpreting $B_{S,t}(m_{L_t}) $  as the speaker's belief about the listener's meaning $  m_{L_t}  $  at turn $t$.

\section{Modeling Conversations Using Pragmatic LLMs}\label{sec:mddial}

%In this section, we provide preliminary evidence that the CRSA model can produce reasonably good estimations of both the likelihood of each utterance $u_t$ and the target $y$ of the task in doctor–patient conversations, which is the disease corresponding to the symptoms described by the patient. 
In this section, we present preliminary evidence that the CRSA model can estimate both utterance likelihoods and task targets in doctor–patient conversations. Specifically, it improves the mean perplexity for conversation utterances and of the final diagnosis prediction, compared to the raw outputs of the LLM. To this end, we used the MDDial dataset~\citep{macherla_mddial_2023}, which consists of template-based conversations between a doctor and a patient. In each dialog, the patient is assigned a subset of predefined symptoms, and the doctor must determine the correct disease from a set of possible pathologies.

\paragraph{Methodology} As anticipated in Section~\ref{sec:crsa_eq}, in order to apply the pragmatic models, we compute the literal speaker with equation~\eqref{eq:literal_speaker} using a pre-trained LLaMA3.2–1B-Instruct\footnote{https://huggingface.co/meta-llama/Llama-3.2-1B-Instruct} language model. In this equation, $\mathrm{prompt}(m_{S_t})$ is the text used to prompt the model with the relevant medical scenario. When $S_t$ is the doctor, the prompt includes specific instructions to ask questions and produce a diagnosis, followed by two example doctor–patient conversations. When $S_t$ is the patient, the prompt instructs the model to play the role of the patient. It uses the same conversation examples as in the doctor prompt but additionally includes the patient’s current symptoms at that turn. The full prompts used can be found in Appendix~\ref{app:prompts}. Importantly, we assume that the set of possible utterances is pre-defined, and we compute the speaker probability over this set. We use the literal speaker as lexicon in Equation~\eqref{eq:literal_listener} for computing the literal listener. 

We compute $P(m_{patient},m_{doctor},y_{diagnose})$ by counting the number of times that symptoms uttered by the patient appear in the context of a certain diagnosis. Note that for this case, the number of possible meanings for the doctor is 1, since it is assumed he/she has allways the same background on knowledge in the field.

% \subsection{Numerical results and discussion}
\paragraph{Metrics} To evaluate performance, we compute the speaker perplexity as
$ 
PPL = \frac{1}{N} \sum_{i=1}^N  \exp\left(-\sum_{t=1}^{T_i} \log S_t\big(u^{(i)}_t \mid w^{(i)}_t, m^{(i)}_{S_t}\big)\right),
$ 
where $N$ is the number of rounds and $T_i$ is the number of turns in round $i$.  For the listener, we compute the task success rate using the listener of the last step $T_i$ of each round
$ 
TSR = -\frac{1}{N} \sum_{i=1}^N \mathbbm{1}_{\{ y^{(i)} \,=\, \argmax_{y\in\yspace} L_{T_i}\big(y \mid u_{T_i}, m_{L_{T_i}}, w_{T_i}\big)\}}
$ ,
which is analogous to the method described in Section~\ref{sec:find_a1}.

\paragraph{Results} The results are presented in Table~\ref{tab:mddial_results} for the train split of the dataset (1878 samples) and for a value of $\alpha=2.5$, which is the same as used for Section~\ref{sec:find_a1}. We observed the same trend mentioned in that Section when varying the value of $\alpha$. The CRSA model achieves best performance in terms of both perplexity and accuracy (task sucess rate) compared to the classic RSA and the Literal models. We noted however that the difference between the classic RSA and CRSA is not very large, possibly due to the fact that for this task the agent playing the role of the doctor struggles to obtain a good estimation of the patient's belief. Since this estimation is close to uniform at all steps, both models converge to nearly identical equations and thus exhibit similar performance. Still, the low task success rate suggests that the setting is inherently difficult and may require more discriminative models. CRSA provides a useful structure in this direction, and it is plausible that coupling it with a stronger LLM could yield a more informative initialization of the speaker and, in turn, improved task performance.

\begin{table}[t]
    \centering
    \begin{tabular}{lrr}
    \hline
     & Speaker PPL & Task success rate \\
    \hline
    CRSA & \textbf{15.362} & \textbf{0.085} \\
    RSA & 15.395 & \textbf{0.085} \\
    Literal & 60.835 & 0.079 \\
    \hline
    \end{tabular}
    \caption{Speaker perplexity (PPL) and Task success rate for $\alpha=2.5$ of the listener and the speaker for each model, computed for the MDDial dataset.}
    \label{tab:mddial_results}
\end{table}

\section{Possible Future Directions of this work}

There are many ways in which the CRSA model can be improved. One of the major limitations of the model is that there is no systematic way of directly modeling the meaning spaces $\mespace_A$ and $\mespace_B$, which are always application-dependent. One possible way of moving towards a scalable application of the CRSA model in the large language model architecture is by modeling these spaces as continuous. That is, producing an embedding representation of each private meaning, $e_{A},e_{B} \in \mathbb{R}^d$, and incorporating that embedding into the computation of $S_t(u_t|w_t,m_{S_t})$. Although the equations of the model remain essentially the same in this scenario, this approach opens many challenging points. For instance, the computation of the sums in the model's equations, which now become integrals; the way of combining the language model with the embedding $e_{S_t}$ in order to compute $S_t(u_t|w_t,m_{S_t})$, which is definitely non-unique; or the modeling of $p(m_A,m_B,y)$, which now becomes a mixed probability function.

In addition to this, there is the problem of modeling the space of utterances, which is inherited from classic RSA. However, since the past utterances are part of the design of the CRSA, the natural way to scale this model to more realistic applications in which generation is done token by token is by directly replacing utterances with tokens. We expect that this shift may influence the model’s pragmatic capabilities, since the reasoning is performed at the token level, not at the utterance level. We intend to investigate these trade-offs carefully in future work.

Finally, there are many ways in which the original gain function from which the equations of the model are derived could be modified depending on the application scenario. For instance, situations in which the meanings are not fixed in time, or where more than two agents participate in the dialog, can also fit within a similar procedure to that used in this work. This allows for the introduction of pragmatic reasoning in more realistic scenarios in the same mathematically grounded way as was done in the current work. 

\section{Summary and Concluding Remarks}

In this work, we introduced the Collaborative Rational Speech Act (CRSA) framework, an information-theoretic extension of RSA tailored for principled pragmatic reasoning in multi-turn, task-oriented dialogs. By integrating a novel multi-turn gain function grounded in interactive rate-distortion theory, CRSA effectively models the evolving belief dynamics of both interlocutors, overcoming key limitations of traditional RSA in collaborative contexts. Our preliminary results demonstrate that CRSA successfully captures the progression of shared understanding, partner beliefs, and utterance generation, providing the way for more natural and efficient communication in complex conversational settings.

%Through experiments on simulated referential games and template-based doctor–patient interactions, we demonstrate that CRSA produces more consistent and collaborative behavior than existing baselines, particularly in contexts that require belief alignment and implicit coordination. These results underscore the potential of information-theoretic pragmatics as a foundation for designing trustworthy and cooperative dialog agents.

CRSA lays the foundation for developing conversational agents driven by mathematically grounded principles of pragmatic reasoning. This principled formulation enhances both the interpretability and controllability of agent behavior, enabling the construction of language models that move beyond surface-level fluency to demonstrate structured, socially coherent, and contextually appropriate dialog. In this way, CRSA represents a significant step toward building pragmatic agents whose interactions are not only effective but also firmly rooted in the formal theory of communication.

\section*{Limitations}

This work focuses on simulated referential games and template-based doctor–patient dialogs, which, while controlled and insightful, do not capture the full variability and complexity of real-world conversations. Additionally, the CRSA framework relies on a fixed, predefined set of possible utterances at each turn, limiting its applicability to open-ended or generative dialog scenarios involving variable-length token sequences. These factors currently restrict the scalability of our approach to more naturalistic domains. Future work will aim to overcome these limitations by extending CRSA to handle dynamically generated utterance spaces and by evaluating its effectiveness in less structured, real-world conversational settings.

\section*{Ethical considerations}

This work presents a theoretically grounded framework for pragmatic reasoning in multi-turn dialogs. It is primarily methodological and does not involve direct deployment or interaction with real users. The datasets employed—simulated referential games and template-based medical dialogs—are synthetic and contain no personal or sensitive data.

However, since CRSA aims to inform the development of more interpretable, goal-driven conversational agents, potential applications in sensitive domains like automatic medical diagnosis raise important ethical considerations. In such contexts, errors in belief tracking or task inference could result in incorrect recommendations, especially if users overestimate the system’s understanding or authority. While our medical domain experiments are purely illustrative and not intended for clinical use, they underscore the critical need for caution when adapting theoretical models to real-world diagnostic settings. Future deployments must involve rigorous domain-specific validation, proper oversight, and human supervision to ensure safety and reliability.

A final potential ethical concern is the risk of anthropomorphizing AI systems when they are described as communicative agents. While the agent metaphor is useful for modeling and analysis, it may inadvertently suggest that such systems possess autonomy, intentionality, or even consciousness. We stress that this is not the case: our use of agent-like terminology is strictly metaphorical and does not imply any deep philosophical claims about the nature of AI systems.

\section*{Acknowledgments}

We thank the anonymous reviewers for their insightful feedback. This work benefited from the resources (GPUs and CPUs) of Lab-AI, an institution member of Université Paris-
Saclay, for running the experiments.

\bibliography{anthology,custom}

\appendix

\section{Detailed Expressions of the YRSA Model}\label{app:yrsa}

In \citet{zaslavsky_ratedistortion_2021}, the authors propose to use the alternation maximization (AM) algorithm~\citep{csiszar_information_2004} to maximize the gain function of expession~\ref{eq:rsa_gain}:
\begin{align*}
    S^{k+1} &= \argmax_{S} \mathcal{G}(S,L^k), \\
    L^{k+1} &= \argmax_{L} \mathcal{G}(S^{k+1},L).
\end{align*}
If the same procedure is applied to the gain of Equation~\eqref{eq:yrsa_gain} (the one corresponding to the YRSA model), then the following equations are obtained:
\begin{align}
    S^{k+1}&(u|m_A) \propto \nonumber \\ &\exp\big[\alpha(\sum_{\forall (m_B,y)}P(m_B,y|m_A) \nonumber \\
    &(\log(L^{k}(y|m_B,u)) - C(u))\big], \\[.5em]
    L^{k+1}&(y|m_B,u) \propto \nonumber \\ 
    &\sum_{\forall m_A}P(m_A,m_B,y)\cdot S^{k+1}(u|m_A).
\end{align}
Additionally, if a lexicon $\lexicon(u,m_A)$ is given, the listener is initialized as
\begin{align}
    L^0(y|m_B,u) &\propto \sum_{\forall m_A}P(m_A,m_B,y)\cdot \lexicon(u,m_A).
\end{align}
The proof of how to arrive to these equations is very similar to the ones to obtain the CRSA, which is presented in appendix~\ref{app:proof} so we suggest to read that Section instead.

\section{Derivation of the CRSA Model Expressions}
\label{app:proof}

For the following derivation we have assumed that the speaker is agent A and the listener is agent B in order to simplify notation. In addition, since every variable depends on the turn, we will omit the subindex $t$ for the same reason. We start by representing the speaker, the listener, the prior and the cost as matrices:
\begin{align*}
     s_{awu} &=S(u|m_A,w) = [\matS]_{awu} \\ \mathbf{S} &\in [0,1]^{\mespace_A\times \wspace \times \uspace} \\[.5em]
    l_{buwy} &= L(y|m_B,u,w) =[\matL]_{buwy} \\ \mathbf{L} &\in [0,1]^{\mespace_B\times \uspace \times \wspace \times \yspace} \\[.5em]
    P_{abyw} &= P_S(m_A,m_b,y,w) = [\matP]_{abyw} \\ \matP &\in [0,1]^{\mespace_A \times \mespace_B \times \yspace \times \wspace} \\[.5em]
    c_{u} &= C(u) = [\matC]_{u} \\ \matC &\in \mathbb{R}^{\uspace}
\end{align*}
with the restrictions
\begin{align}\label{eq:restrictions}
    \sum_u s_{awu} = 1,  & &
    \sum_y l_{buwy} = 1.
\end{align}
The gain function at the turn $t$ as a function of the matrices $\matS$ and $\matL$ can be written as
\begin{align}
    \mathcal{G}(\matS,\matL)&= 
    -\sum_{abywu}s_{awu}P_{abyw} (\log s_{awu} + \nonumber \\ & \alpha (\log l_{buwy} - c_u)) \nonumber \\[.5em]
    &=-\sum_{awu}s_{awu}P_{aw}\log s_{awu} + \nonumber \\ 
    & \alpha \sum_{abywu} s_{awu}P_{abyw} \log l_{buwy} - c_u \nonumber \\
    &=\sum_w \mathcal{G}_w(\matS,\matL),
\end{align}
where
\begin{align}
    \mathcal{G}_w(\matS,\matL) =-\sum_{au}s_{awu}P_{aw}\log s_{awu} \nonumber \\ 
    + \alpha \sum_{abyu} s_{awu}P_{abyw} \log l_{buwy} - c_u. 
\end{align}
Since the overall gain is a sum of the gain for a specific utterance history $w$, taking the derivative with respect to a different value of $w$ cancels out the other terms in the sum, so we can abbreviate the notation by omitting the $w$ subindex. Then, the problem reduces to maximize the following Lagrangian:
\begin{align*}
    \mathcal{L}(\matS,\matL) &= -\sum_{au}s_{au}P_{a}\log s_{au} \\ &+ \alpha \left(\sum_{abyu} s_{au}P_{aby} \log l_{buy} - c_u \right) \nonumber \\
    &- \sum_a \lambda_{a} g_a(\matS) - \sum_{bu}\lambda_{bu} g_{bu}(\matL)
\end{align*}
with
\begin{align*}
    g_{a}(\matS) &= 1 - \sum_u s_{au} = 0, \\
    g_{bu}(\matL) &= 1 - \sum_y l_{buy} = 0.
\end{align*}
Taking the gradient w.r.t $s_{\hat a \hat u}$ and $l_{\hat b \hat u \hat y}$, we get
\begin{align*}
    \frac{\partial \mathcal{L}}{\partial s_{\hat a \hat u}} &= -P_a(\log s_{\hat a \hat u} + 1) \nonumber \\
    &+ \alpha \sum_{by} P_{aby}(\log l_{b \hat u y} - c_{\hat u}) - \lambda_{\hat a} = 0, \\
    \frac{\partial \mathcal{L}}{\partial l_{\hat b \hat u \hat y}} &= \frac{\alpha}{l_{\hat b \hat u \hat y}} \sum_a s_{a \hat u}P_{a \hat b \hat y} - \lambda_{\hat b \hat u} = 0.
\end{align*}
So it is straightforward to see that
\begin{align*}
    l_{\hat b \hat u \hat y} &\propto \sum_a s_{a \hat u}P_{a \hat b \hat y} \\
    s_{\hat a \hat u} &\propto \exp\left(\alpha \sum_{by} \frac{P_{\hat aby}}{P_{\hat a}}(\log l_{b \hat u y} - c_{\hat u}) \right).
\end{align*}
We can rewrite these equations in terms of the the original probabilities adding the past $w$ and the turn $t$ subindex:
\begin{align*}
    L(y|&m_B,u_t,w_t) \propto \nonumber \\ &\sum_{\forall m_A}S(u_t|m_A,w_t)P_S(m_A,m_B,y,w_t) \\
    S(u_t&|m_A,w_t) \propto \nonumber \\
    &\exp(\alpha \sum_{\forall (m_B,y)} P_S(m_B,y|m_A,w_t) \nonumber \\ 
    &(\log L(y|m_B,u_t,w_t) - C(u_t)).
\end{align*}
Then, by applying equations \ref{eq:belief_decomposition} and \ref{eq:belief_prime} of Section~\ref{sec:crsa} we can directly obtain
\begin{align}
    & S_{t}(u_t|w_t,m_{S_t}) \propto  \\ &\exp(\alpha\sum_{\forall (m_{L_t},y)}B'_t(m_{S_t},m_{L_t},y)V_L(u_t,w_t,m_{L_t},y), \nonumber  \\[.5em]
    & L_t(y|u_t,w_t,m_{L_t}) \propto  \\
    &\sum_{\forall m_{S_t}}B_{S,t}(m_{S_t})P_t(m_{S_t},m_{L_t},y) S_t(u_t|w_t,m_{S_t}). \nonumber
\end{align}
These are the equations that maximize the gain $\mathcal{G}(\matS,\matL)$ subject to the restrictions~\ref{eq:restrictions}. Then, by applying again the alternation maximization algorithm we obtain the CRSA algorithm.

\section{Prompts used in the MDDial dataset}\label{app:prompts}
We prompt two different models for generating the lexicons defined in Section~\ref{sec:mddial}. The first one (the one for the patient) contained the following instructions:

\vspace{1em}

\begin{verbatim}  
You are an assistant that simulates to be 
a patient who has a disease and describes 
the symptoms to the user, which is a 
medical doctor.

Here is an example of a conversation 
between the assitant (i.e., the patient) 
and the user (i.e., the doctor). You are 
experiencing the following symptoms:
Acid reflux, Stomach ache, Bloating
Assistant: Hi Doctor, I am having Acid 
reflux
User: In that case, do you have any 
Stomach ache?
Assistant: Yes most of the times
User: In that case, do you have any 
Bloating?
Assistant: Yes most of the times
User: This could probably be Esophagitis.

Here is an example of a conversation 
between the assitant (i.e., the patient) 
and the user (i.e., the doctor). You are 
experiencing the following symptoms:
Eye swelling
Assistant: Recently, I am experiencing 
Eye swelling
User: Is it? Then do you experience Cry?
Assistant: No, I never had anything like 
that.
User: Oh, do you have any Fever?
Assistant: No, I never had anything like 
that.
User: Oh, do you have any Photophobia?
Assistant: No, I don't have that
User: Ok, this means you might be having 
Conjunctivitis.

Now, participate in a real conversation 
with the user. You are experiencing the 
following symptoms:
{patient symptoms}
\end{verbatim}

The prompt used for the doctor contained the following instructions:

\begin{verbatim}
You are an assistant that simulates to be 
a doctor who is diagnosing a patient based 
on the symptoms that he or she describes. 
You can ask questions to the patient, but 
ultimately, you have to provide a diagnosis 
based on the symptoms described by the 
patient.

Here is an example of a conversation 
between the assitant (i.e., the doctor) 
and the user (i.e., the patient). The 
patient is experiencing the following 
symptoms:
User: Hi Doctor, I am having Acid reflux
Assistant: In that case, do you have 
any Stomach ache?
User: Yes most of the times
Assistant: In that case, do you have 
any Bloating?
User: Yes most of the times
Assistant: This could probably be 
Esophagitis.

Here is an example of a conversation 
between the assitant (i.e., the doctor) 
and the user (i.e., the patient). The 
patient is experiencing the following 
symptoms:
User: Recently, I am experiencing Eye 
swelling
Assistant: Is it? Then do you 
experience Cry?
User: No, I never had anything like 
that.
Assistant: Oh, do you have any Fever?
User: No, I never had anything like 
that.
Assistant: Oh, do you have any 
Photophobia?
User: No, I don't have that
Assistant: Ok, this means you might 
be having Conjunctivitis.

Now, participate in a real conversation 
with the user. You can ask questions to 
the patient, but ultimately, you have 
to provide a diagnosis based on the 
symptoms described by the patient.
\end{verbatim}

\section{Errors intervals in the reference game}
\begin{figure}
    \centering
    \includegraphics[width=\linewidth]{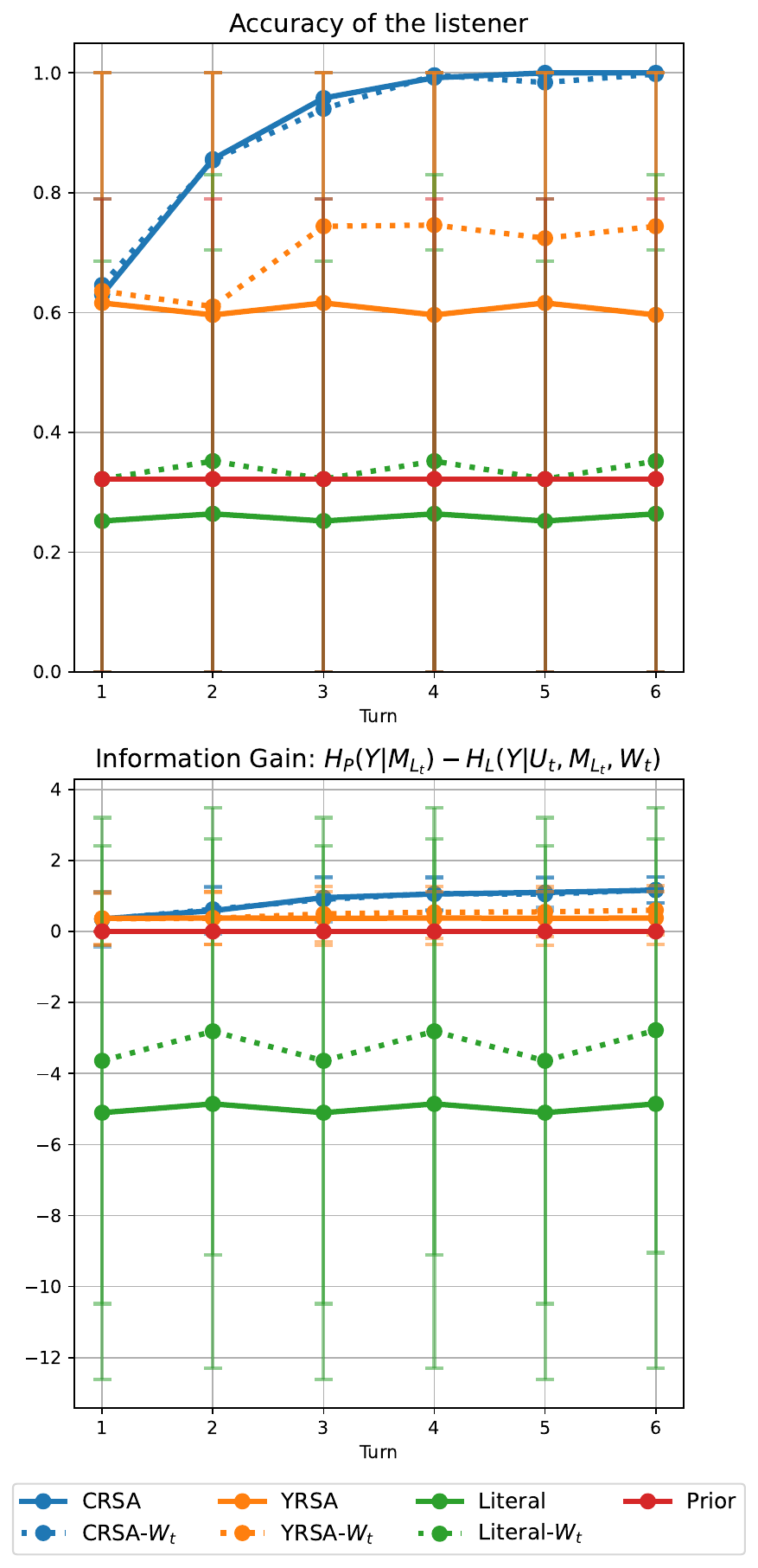}
    \caption{Performance of the CRSA compared to baselines, including error bars.}
    \label{fig:find_a1_preformance_err}
\end{figure}

Figure~\ref{fig:find_a1_preformance_err} shows the same results as Figure~\ref{fig:find_a1_performance} but with the standard deviation of each model. We did not include this plot in the main text for readability, but it can also be noted that the CRSA reduces the variance of the results in comparison with the other models.

\end{document}